\documentclass[runningheads]{llncs}

\usepackage{eccv}

\usepackage{eccvabbrv}

\usepackage{graphicx}
\usepackage{booktabs}

\usepackage[accsupp]{axessibility}  % Improves PDF readability for those with disabilities.

\usepackage{hyperref}

\usepackage{orcidlink}

\usepackage{multirow}
\usepackage{amssymb}

\usepackage{wrapfig}

\begin{document}


\title{Sparse-Aware Vector Quantization for Bandwidth-Efficient Collaborative 3D Semantic Occupancy Prediction} 

\titlerunning{VQSOP}

\author{Feng Li\inst{1} \and
Chaokun Zhang\inst{2}\thanks{Corresponding author.} \and
Gong Chen\inst{1}}

\authorrunning{F. Li et al.}

\institute{School of Computer Science and Technology, Tianjin University, Tianjin, China \and
School of Cybersecurity, Tianjin University, Tianjin, China \\
\email{\{fengli0116, zhangchaokun, gongchen01\}@tju.edu.cn}}
\maketitle

\begin{abstract}
Collaborative perception extends single-agent perception by enabling multiple vehicles to exchange complementary perceptual information. However, it introduces an inherent trade-off between perception gain and communication overhead, which is particularly severe for 3D semantic occupancy prediction that relies on fine-grained spatial structures. Existing methods typically compress 3D features into 2D, causing severe spatial information loss, or transmit dense 3D representations, hindering real-world deployment. To overcome these limitations, we propose a bandwidth-efficient collaborative \textbf{V}ector \textbf{Q}uantization \textbf{S}emantic \textbf{O}ccupancy \textbf{P}rediction (VQSOP) framework. VQSOP employs a Sparse-Aware Vector Quantization (SAVQ) mechanism that exploits 3D scene sparsity to compactly encode informative regions, drastically reducing communication overhead while preserving complete geometric context. Furthermore, to enhance structural consistency and feature continuity, we design a Dual-Branch Adaptive Spatial Refinement (ASR) module that dynamically fuses local high-frequency details with broad contextual semantics. Extensive experiments demonstrate that our approach achieves state-of-the-art performance while reducing communication volume by up to 82$\times$.
  \keywords{Collaborative perception \and 3D occupancy prediction \and Autonomous driving}
\end{abstract}

\begin{figure}[tb]
  \centering
  \includegraphics[width=\linewidth]{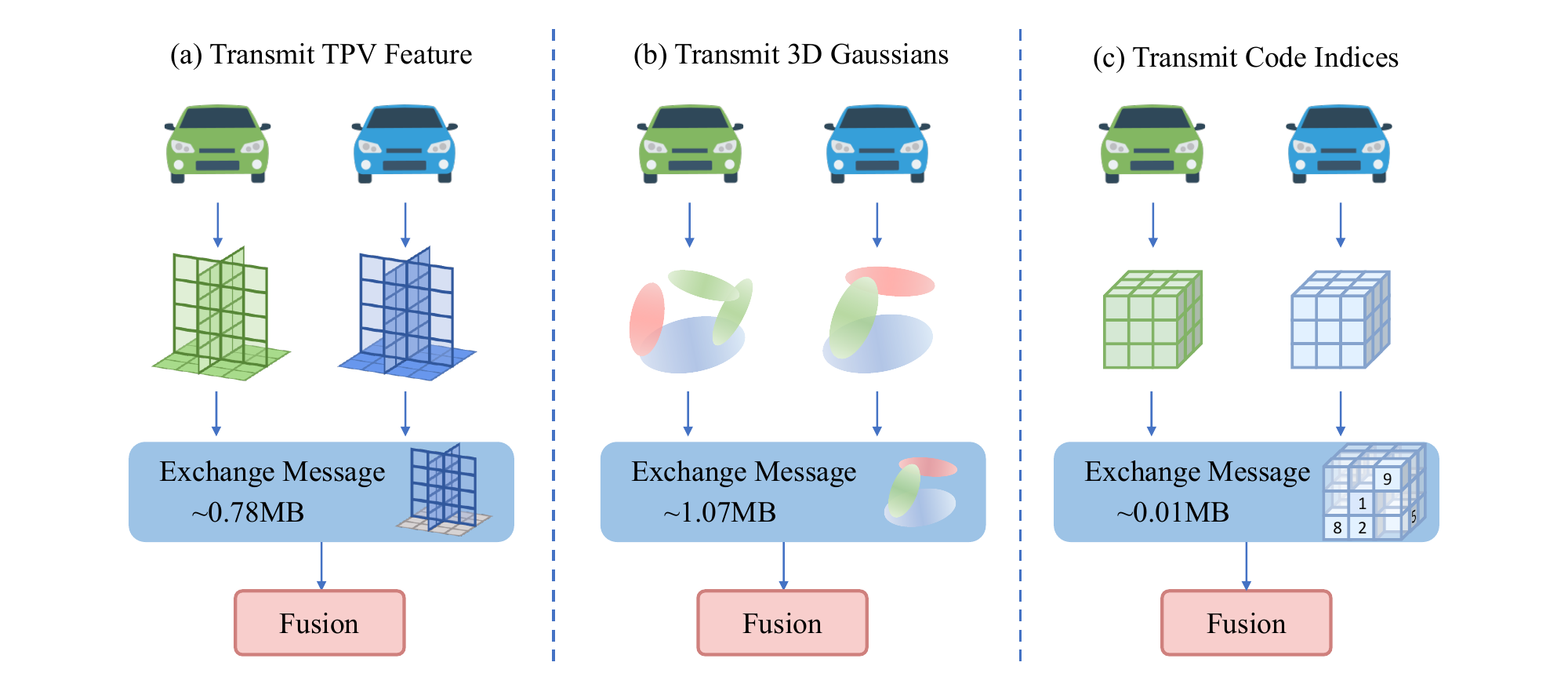}
  \caption{
    Comparison of shared feature representations. (a) Transmitting TPV features loses geometric details and complicates spatial alignment. (b) Transmitting 3D Gaussians tightly couples prediction performance with communication bandwidth, as the number of Gaussians directly impacts accuracy. (c) Our method transmits more critical perception information via compact code index messages, drastically reducing the communication volume while preserving essential semantic and geometric information for effective collaborative fusion.
  }
  \label{fig:introduction}
\end{figure}
\section{Introduction}
\label{sec:intro}

Reliable environmental perception forms the foundation of autonomous driving systems, as it directly determines the safety and effectiveness of downstream planning and control 
modules \cite{omeiza2022explanations,pradeep2023reliability,yang2023aide}. 
By leveraging Vehicle-to-Everything (V2X) communication, collaborative perception enables multiple agents to exchange complementary information and extend the effective perception range \cite{xu2022v2x,hu2023collaboration,yang2023spatio,hu2022where2comm}. This paradigm overcomes the inherent limitations of traditional single-agent perception systems caused by restricted sensor fields of view and severe occlusions \cite{gao2024survey,yang2023aide,xiang2023hm}, thereby significantly enhancing robustness in complex and dynamic traffic environments.

Previous collaborative perception research predominantly focused on 3D object detection and Bird's-Eye-View (BEV) segmentation tasks \cite{cui2022coopernaut,wang2024deepaccident,li2021learning,wang2020v2vnet,tan2024dynamic,xu2022cobevt}. These conventional tasks simplify scene perception by omitting crucial 3D semantic details. Furthermore, BEV-based features inherently suffer from information loss due to height compression. Such limitations restrict the ability to identify irregularly shaped or heavily occluded objects, ultimately hindering holistic scene understanding and downstream decision-making. Recently, 3D semantic occupancy prediction has garnered significant attention for its ability to provide detailed 3D semantic and geometric information by utilizing fine-grained voxel-based representations \cite{cao2022monoscene,wei2023surroundocc,huang2023tri,huang2024gaussianformer,li2024viewformer,duan2025sdgocc}. 

Despite its tremendous potential to offer a richer understanding of the environment, 3D semantic occupancy prediction remains largely under-explored in collaborative settings \cite{wu2025synthetic}. Fundamentally, collaborative perception involves an inherent trade-off between perception gain and communication overhead. This conflict is particularly severe for 3D occupancy prediction, where high-dimensional spatial structures make the direct transmission of dense features impractical under limited communication bandwidth. CoHFF \cite{song2024collaborative} proposed an early collaborative semantic occupancy prediction framework, demonstrating the immense potential of V2X feature fusion for semantic occupancy prediction. However, CoHFF adopts a tri-perspective view (TPV) representation that relies on explicit depth supervision to preserve spatial consistency. Furthermore, to reduce communication volume, it only transmits orthogonal plane features, resulting in inevitable information loss (\cref{fig:introduction}(a)). In an effort to retain complete 3D structural details without relying on 2D projections, recent methods have attempted to introduce 3D Gaussians \cite{chen2026vision}. Although these approaches preserve richer geometric expressiveness, prediction accuracy directly depends on the number of transmitted Gaussians. This tightly couples performance with communication bandwidth, fundamentally restricting their practical deployment in bandwidth-constrained collaborative perception systems (\cref{fig:introduction}(b)).

In this paper, we propose a novel collaborative Vector Quantization Semantic Occupancy Prediction (VQSOP) framework that significantly improves the perception-communication trade-off in multi-agent systems. VQSOP introduces a Sparse-Aware Vector Quantization (SAVQ) mechanism that exploits inherent spatial sparsity and local feature redundancy in autonomous driving scenarios to selectively focus on informative regions. Instead of transmitting dense continuous features, this mechanism directly quantizes the selected high-dimensional feature representations into compact discrete codes defined in a shared codebook, transmitting only the corresponding indices. This quantization-driven design effectively reduces communication overhead while preserving essential 3D geometric context. Furthermore, to enhance semantic perception capabilities, we design a dual-branch Adaptive Spatial Refinement (ASR) module. The module consists of a local branch for preserving fine-grained geometric details and a context branch for modeling long-range semantic dependencies. These complementary features are adaptively fused via spatially adaptive weighting, enabling the model to reconstruct coherent and semantically consistent volumetric representations with high fidelity (\cref{fig:introduction}(c)).

Extensive experiments validate the effectiveness of our proposed approach. Specifically, under the single-agent setting, VQSOP outperforms the strongest baseline by 4.41\% in mIoU and 2.64\% in IoU. Furthermore, in the collaborative setting, it surpasses state-of-the-art methods by 4.10\% and 0.92\% in mIoU and IoU, respectively. Notably, VQSOP reduces communication volume by up to 82$\times$ while delivering superior collaborative perception performance. 

Our main contributions can be summarized as follows:

\begin{itemize}
\item We propose VQSOP, a novel collaborative semantic occupancy prediction framework. It compresses features into discrete code indices for transmission, effectively preserving perception-relevant information.
\item We design a SAVQ mechanism to compactly encode key informative regions. Additionally, we introduce an ASR module to dynamically aggregate spatial features and enhance the overall semantic perception capabilities.
\item We conduct comprehensive experiments to validate our approach, demonstrating that VQSOP achieves state-of-the-art performance and establishes a highly superior perception-communication trade-off.
\end{itemize}

The code is available at \url{https://github.com/cheerfulli/VQSOP}.

\section{Related Work}

\subsection{3D Semantic Occupancy Prediction}
3D semantic occupancy prediction has attracted growing interest in recent years, as it provides a holistic representation of driving environments by predicting both occupancy and semantic labels for all voxels within a predefined spatial range. Existing methods can be broadly categorized based on their input sensory modalities. LiDAR-based approaches leverage the precise geometric measurements of 3D point clouds to generate high-quality voxel predictions \cite{cheng2021s3cnet,yan2021sparse,zuo2023pointocc}. Meanwhile, vision-centric approaches have dominated the mainstream due to their rich semantic context and remarkable cost-effectiveness.  
SurroundOcc \cite{wei2023surroundocc} utilized spatial attention to lift 2D features into 3D space in a multi-scale manner, employing multi-frame point clouds to construct dense occupancy ground truths. TPVFormer \cite{huang2023tri} proposed a tri-perspective view representation for describing 3D scenes in semantic occupancy prediction. OccFormer \cite{zhang2023occformer} designed a dual-path Transformer network to process 3D volume features efficiently. SGN \cite{mei2024camera} presented a dense–sparse–dense architecture that adaptively identifies sparse seed voxels and incorporates hybrid guidance to promote the convergence of semantic propagation. GaussianFormer \cite{huang2024gaussianformer} introduced an object-centric approach that describes 3D scenes using sparse 3D semantic Gaussians. 
Liu et al. \cite{liu2024fully} proposed a fully sparse occupancy network that reconstructs sparse geometry with a sparse voxel decoder and predicts semantic occupancy using sparse queries. Tang et al. \cite{tang2024sparseocc} leveraged a lossless sparse latent representation to achieve efficient semantic occupancy prediction. OPUS \cite{wang2024opus} formulates occupancy prediction as a direct set prediction problem, leveraging learnable queries to predict occupied locations and classes.
Despite the remarkable progress achieved by these vision-centric semantic occupancy prediction methods, they are primarily designed for single-agent systems. Directly extending these high-dimensional 3D representations to multi-agent collaborative scenarios introduces severe communication bottlenecks.

\subsection{Collaborative Perception}
Collaborative perception paradigms can be broadly classified into early, late, and intermediate fusion. Early fusion methods \cite{chen2019cooper,su2024collaborative} share raw sensor data between connected agents to expand the perceptive field, but consume excessive communication bandwidth, severely limiting their applicability in real-time. Late fusion approaches \cite{arnold2022cooperative,shi2022vips} share and fuse the final detection outputs from individual vehicles. Although this paradigm is highly bandwidth-efficient, it is extremely sensitive to the individual detection performance of each agent, where a single localized error can easily degrade the final consensus. To strike an optimal balance between communication overhead and perception accuracy, recent efforts have predominantly focused on intermediate fusion, where agents share processed intermediate feature representations. V2X-ViT \cite{xu2022v2x} introduces a heterogeneous multi-agent attention module to fuse information from vehicles and infrastructure. Where2comm \cite{hu2022where2comm} proposes a novel spatial confidence-aware communication strategy that focuses on key perceptual regions, enabling performance improvement with reduced communication overhead. CORE \cite{wang2023core} utilizes LiDAR measurements and adopts spatial-channel compression along with attention-based collaboration for efficient scene reconstruction. CoBEVT \cite{xu2022cobevt} designs a local-global sparse attention mechanism to capture complex spatial interactions across different views and agents. CoBEVFusion \cite{qiao2024cobevfusion} investigates multi-modal fusion by introducing a Dual Window-based Cross-Attention module, which integrates camera and LiDAR features through CNN-based inter-agent interaction.
WhisperNet \cite{chen2026whispernet} proposes a novel receiver-centric global coordination paradigm that jointly optimizes feature transmission in the spatial and channel dimensions across agents.
However, these intermediate fusion strategies are predominantly tailored for 3D object detection or 2D BEV segmentation. Directly applying them to transmit dense, high-dimensional features for 3D semantic occupancy prediction incurs an unacceptable bandwidth burden.

\section{Method}

\subsection{Problem Formulation}

We model the collaborative perception system as a communication graph $\mathbb{G} = (\mathcal{A}, \mathcal{E})$, where $\mathcal{A}$ is the set of agents, and $\mathcal{E}$ represents the existing communication links between two agents. For an agent $i \in \mathcal{A}$, its connected neighbors are defined as $\Omega_i = \{j \mid (i, j) \in \mathcal{E}, j \neq i\}$. Let $\mathbf{X}_i$ denote the multi-view RGB images captured by the surround cameras mounted on agent $i$. The holistic surrounding environment is represented as 3D voxels with one-hot embeddings $\mathbf{O} \in \mathbb{R}^{X \times Y \times Z \times C}$, where $X, Y$, and $Z$ represent the spatial dimensions, and $C$ is the number of semantic categories. For each agent $i$, $\mathbf{O}_i$ denotes the predicted occupancy of the voxels, while $\mathbf{O}_i^*$ represents the corresponding ground truth. The collaborative semantic occupancy prediction task is formulated as a constrained optimization problem:
\begin{equation}\label{eq:opt}
\begin{split}
    \max_{\theta, \mathcal{P}} & \sum_{i \in \mathcal{A}} g \Big( \Phi_\theta \big( \mathbf{X}_i, \{ \mathcal{P}_{j \to i} \}_{j \in \Omega_i} \big), \mathbf{O}_i^{*} \Big) , \\
    & \text{s.t.} \quad \sum_{i \in \mathcal{A}} \sum_{j \in \Omega_i} b(\mathcal{P}_{j \to i}) \leq B,
\end{split}
\end{equation}
where $\Phi_{\theta}$ denotes the collaborative perception network parameterized by $\theta$, and $g(\cdot)$ is the prediction evaluation metric. $\mathcal{P}_{j \to i}$ is the message transmitted from agent $j$ to agent $i$, and $b(\cdot)$ measures the communication cost of the collaborative messages. Ultimately, the optimization objective is to maximize the overall perception effectiveness under the communication upper bound $B \in \mathbb{R}^+$.

\begin{figure}[tb]
  \centering
  \includegraphics[width=\linewidth]{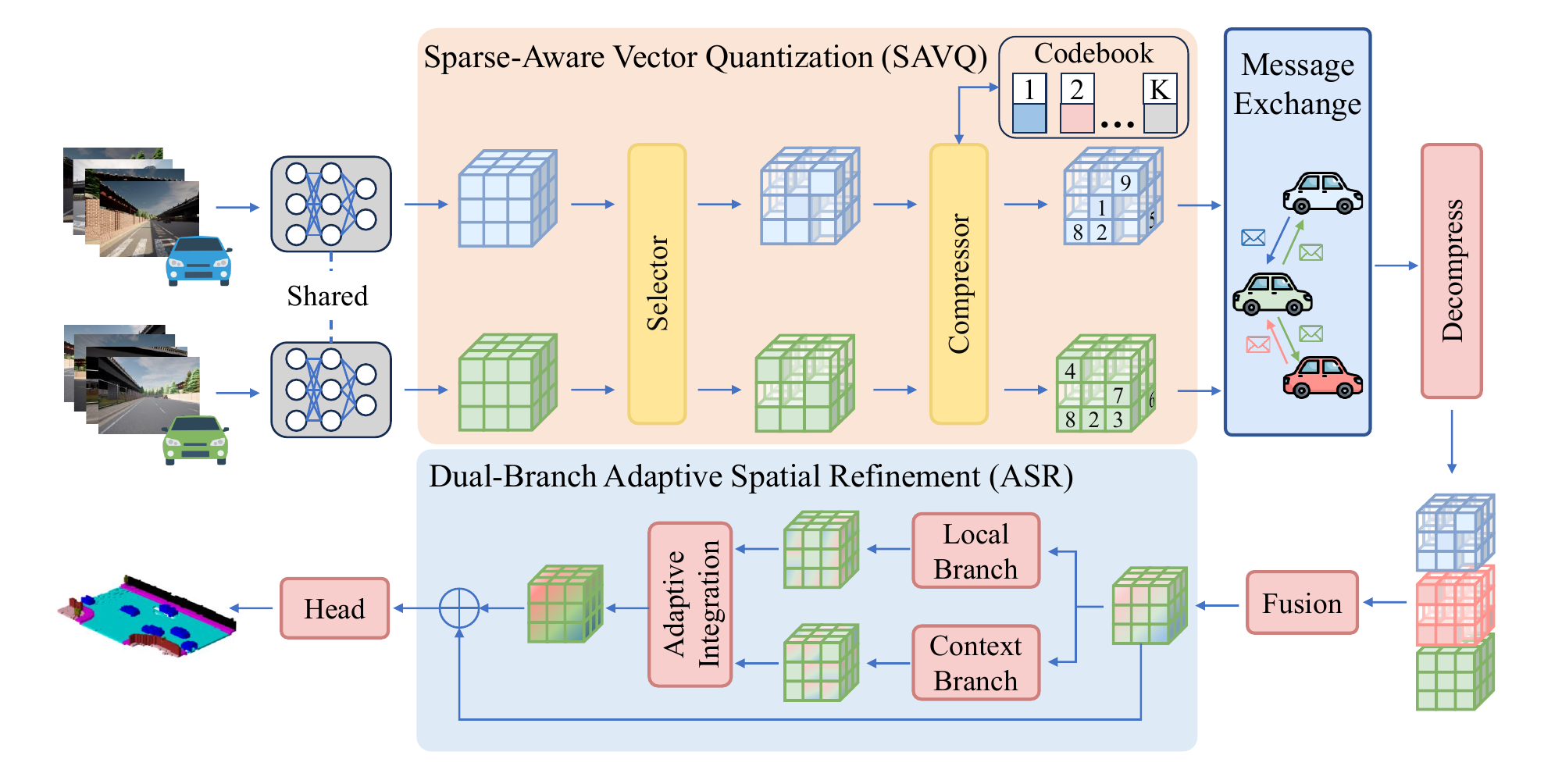}
  \caption{
  Overall architecture of the proposed VQSOP framework. The pipeline consists of three main stages: (1) SAVQ mechanism, which compresses dense 3D spatial features into discrete code indices for bandwidth-efficient transmission; (2) message decompression and fusion, where received neighbor messages are reconstructed and spatially aggregated with the ego agent's local representation; and (3) ASR module, which dynamically integrates local geometric details and broad contextual semantics via spatially adaptive weighting to yield the final 3D semantic occupancy prediction.
  }
  \label{fig:overall}
\end{figure}

\subsection{Overall Architecture}
\cref{fig:overall} illustrates the overall pipeline of our proposed VQSOP framework. Each agent utilizes a shared backbone to process multi-view RGB images and metadata, lifting them into continuous 3D occupancy features. To overcome communication bandwidth bottlenecks, these local features are processed by the SAVQ mechanism before transmission. Specifically, a selector isolates informative spatial regions, and a compressor maps these continuous features into highly compact discrete indices by querying a learnable codebook. These discrete indices are then transmitted through the V2X message exchange network. Upon reception, the message decompression module queries the shared codebook to reconstruct the continuous 3D neighbor features from the received indices. Subsequently, the collaborative fusion module spatially aggregates these reconstructed features with the ego agent's local representation. To further enhance the semantic perception capabilities, the fused features are fed into the ASR module. The ASR utilizes a local branch to extract fine-grained geometric details and a context branch to capture broad contextual semantics. These diverse spatial features are dynamically integrated via spatially adaptive weighting and a residual connection to reconstruct high-fidelity structural details. Finally, the refined features are passed to a task-specific prediction head to generate the ultimate 3D semantic occupancy prediction.

\subsection{Sparse-Aware Vector Quantization}
\begin{wrapfigure}[18]{r}{0.40\textwidth}
  \centering
  \includegraphics[width=\linewidth]{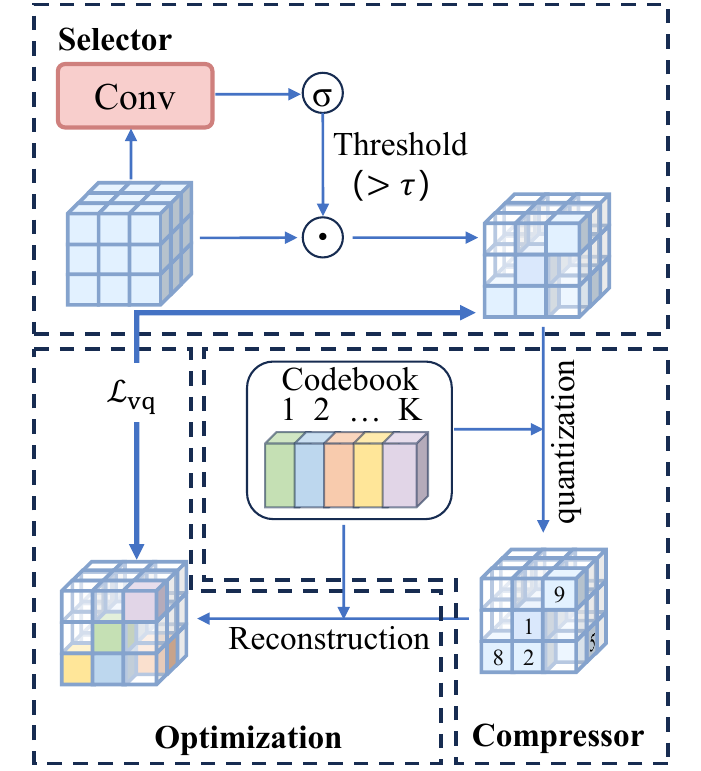} 
  \caption{Architecture of the SAVQ mechanism.}
  \label{fig:SAVQ}
\end{wrapfigure}In collaborative perception, 3D voxel features require higher communication bandwidth than 2D BEV representations, making the direct transmission of dense 3D volumes a major bottleneck for real-world deployment. Nevertheless, we observe that 3D driving scenes naturally exhibit high spatial sparsity, where the vast majority of voxels are empty, implying that only a small fraction contains informative semantic and geometric cues. Therefore, to effectively filter key regions and compress high-dimensional features, we design the SAVQ mechanism. As shown in \cref{fig:SAVQ}, this mechanism selectively retains informative regions and maps the continuous features into a compact discrete codebook representation, thereby significantly reducing communication costs.

\textbf{Sparse-Aware Selector.} For the $j$-th agent in the collaborative network, given a dense 3D feature volume $\mathbf{V}_j \in \mathbb{R}^{X \times Y \times Z \times C}$ extracted by the shared backbone, we employ a lightweight convolutional module to predict a spatial confidence map $\mathbf{S}_j \in [0, 1]^{X \times Y \times Z}$, which indicates the importance of each voxel. A binary sparse mask $\mathbf{M}_j$ is then generated by applying a predefined confidence threshold $\tau$:
\begin{equation}\label{eq:mask}
    \mathbf{M}_j(x,y,z) = 
    \begin{cases} 
        1, & \text{if } \mathbf{S}_j(x,y,z) > \tau, \\ 
        0, & \text{otherwise.} 
    \end{cases}
\end{equation}

By applying this mask to the original feature volume, we obtain the filtered feature map $\mathbf{F}_j$:
\begin{equation}
    \mathbf{F}_j = \mathbf{V}_j \odot \mathbf{M}_j,
\end{equation}
where $\odot$ denotes element-wise multiplication broadcasted along the channel dimension. This step ensures that the subsequent quantization and transmission are strictly focused on geometry-aware and semantic-rich areas.

\textbf{Codebook-Based Compressor.} To efficiently compress high-dimensional voxel semantic data, we introduce a voxel semantic compression scheme using learnable codebooks. Inspired by \cite{van2017neural,hu2024communication}, we approximate high-dimensional feature vectors using the most relevant codes from a task-driven codebook. Consequently, only integer code indices need to be transmitted, rather than complete feature vectors composed of floating-point numbers. This codebook consists of a set of prototype codes that are optimized to effectively approximate the diverse perceptual features present in driving scenes. Let $\mathcal{C} = \{\mathbf{c}_1, \mathbf{c}_2, \dots, \mathbf{c}_K\}$ denote the learnable codebook, where $K$ is the dictionary size and $\mathbf{c}_k$ is the $k$-th prototype code. For each valid spatial position $(x,y,z)$ selected by the mask (i.e., $\mathbf{M}_j(x,y,z) = 1$), we map the continuous feature $\mathbf{F}_j(x,y,z)$ to a discrete code index via a nearest-neighbor lookup:
\begin{equation}\label{eq:compress}
    I_j(x,y,z) = \arg\min_{k \in \{1, 2, \dots, K\}} \|\mathbf{F}_j(x,y,z) - \mathbf{c}_k\|_2.
\end{equation}

During transmission, the $j$-th agent only needs to broadcast the discrete indices $I_j$ corresponding to the critical regions. The bandwidth required to transmit an index is merely $\log_2(K)$ bits (e.g., 8 bits for $K=256$), which achieves a massive compression ratio compared to the original dense volume.

\textbf{Codebook Optimization.} To ensure the learnable codebook accurately represents the continuous feature space, we optimize it by minimizing the reconstruction error. Let $\hat{\mathbf{F}}_j$ denote the quantized sparse feature map, where each valid position is mapped back to its corresponding prototype: $\hat{\mathbf{F}}_j(x,y,z) = \mathbf{c}_{I_j(x,y,z)}$, and the filtered empty regions remain zero. To optimize the codebook, we define the quantization loss across all agents as:
\begin{equation}\label{eq:vqloss}
    \mathcal{L}_{vq} = \sum_{j \in \mathcal{A}} \sum_{x,y,z} \left\| \mathbf{F}_j(x,y,z) - \hat{\mathbf{F}}_j(x,y,z) \right\|_2^2.
\end{equation}

\subsection{Message Decompression and Collaborative Fusion}
Upon receiving the compact messages from neighboring agents, the ego agent must decode these discrete representations and spatially aggregate them to construct a holistic 3D scene representation.

\textbf{Message Decompression.} 
Given the transmitted integer indices from a neighboring agent $j$, the ego agent reconstructs the quantized feature volume by querying the shared learnable codebook $\mathcal{C}$. Each discrete index corresponds to a prototype vector in the codebook, enabling the recovery of continuous feature representations from compact messages. The reconstructed volume is denoted as 
$\hat{\mathbf{F}}_j \in \mathbb{R}^{X \times Y \times Z \times C}$. 
For each valid spatial location indicated by $I_j(x,y,z)$, 
the feature is obtained through a direct lookup operation,
$\hat{\mathbf{F}}_j(x,y,z) = \mathbf{c}_{I_j(x,y,z)}$.
The remaining background regions that were pruned during transmission are padded with zeros to maintain a consistent spatial tensor structure. 
The resulting feature volume is then spatially aligned and fused with the ego vehicle's local representation in subsequent stages.

\textbf{Collaborative Fusion.} To construct a holistic 3D scene representation, the ego agent aggregates its own local dense feature volume $\mathbf{V}_{i}$ with all reconstructed neighbor features to generate the comprehensive collaborative representation:
\begin{equation}
    \hat{\mathbf{F}}_{fused} = \Psi_{fuse} \Big( \mathbf{V}_{i}, \{ \hat{\mathbf{F}}_j \}_{j \in \Omega_i} \Big),
\end{equation}
where $\Psi_{fuse}(\cdot)$ denotes the feature fusion operator (e.g., an attention-based spatial aggregator) that effectively integrates multi-agent perspectives. This collaborative fused volume $\hat{\mathbf{F}}_{fused}$ subsequently serves as the direct input to the downstream refinement module.

\subsection{Dual-Branch Adaptive Spatial Refinement}
\begin{figure}[tb]
  \centering
  \includegraphics[width=\linewidth]{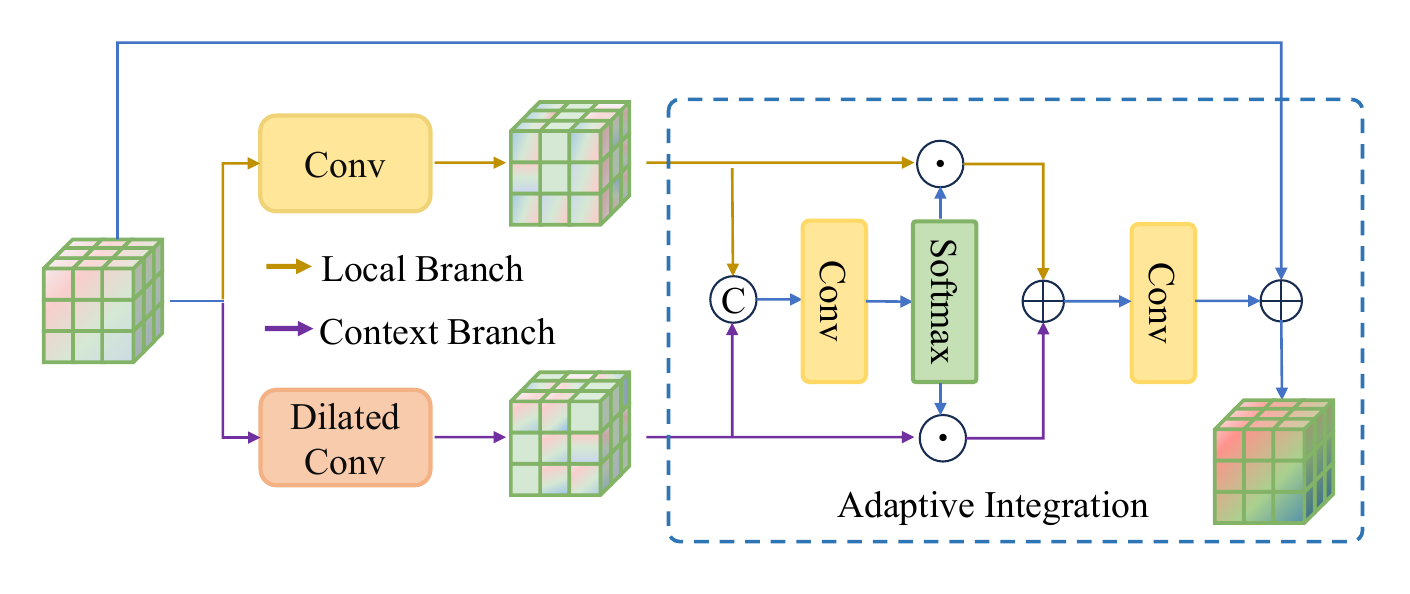} 
  \caption{Architecture of the ASR module. It dynamically aggregates local geometric details and broad contextual semantics through a parallel dual-branch design with spatially adaptive weighting.}
  \label{fig:ASR}
\end{figure}

While collaborative feature fusion enhances spatial awareness, it may blur fine-grained geometric boundaries during aggregation, and long-range contextual dependencies are not fully captured by standard convolutions, hindering precise semantic occupancy prediction.
Therefore, we propose an ASR module to enhance both local geometric details and broad contextual semantics. As shown in \cref{fig:ASR}, the ASR module adopts a parallel dual-branch design: a local branch with standard 3D convolutions for fine-scale details, and a context branch with dilated convolutions for broader dependencies. Their outputs are fused via spatially adaptive weighting for dynamic refinement.

\textbf{Local Branch.} 
The local branch focuses on enhancing fine-grained geometric structures within the 3D feature volume. Specifically, it employs stacked standard 3D convolutional layers with small receptive fields to enhance local spatial continuity and boundary sharpness. This helps recover detailed geometric patterns that may be weakened during multi-agent aggregation, ultimately generating the local representation $\mathbf{F}_{\text{local}} \in \mathbb{R}^{X \times Y \times Z \times C}$.

\textbf{Context Branch.} The context branch aims to capture long-range semantic dependencies across the scene. Specifically, it employs 3D dilated convolutions to efficiently enlarge the spatial receptive field, which enables the aggregation of broader contextual information without sacrificing resolution, facilitates consistent semantic reasoning over spatially distant regions, and produces the context representation $\mathbf{F}_{\text{context}} \in \mathbb{R}^{X \times Y \times Z \times C}$.

\textbf{Adaptive Integration.} Rather than using a naive addition to combine the multi-scale features, we design a spatially adaptive integration mechanism to dynamically fuse the two branches. We concatenate $\mathbf{F}_{\text{local}}$ and $\mathbf{F}_{\text{context}}$ and pass them through a lightweight transformation followed by a Softmax activation. This generates two complementary spatial weight maps, $\mathbf{W}_{\text{local}}, \mathbf{W}_{\text{context}} \in \mathbb{R}^{X \times Y \times Z \times 1}$, which represent the specific demand of each voxel for either boundary details or contextual inpainting:
\begin{equation}
    [\mathbf{W}_{\text{local}}, \mathbf{W}_{\text{context}}] = \text{Softmax}\Big(\text{Conv}_{1\times1\times1}\big(\mathbf{F}_{\text{local}} \parallel \mathbf{F}_{\text{context}}\big)\Big),
\end{equation}

where $\parallel$ denotes channel-wise concatenation. The sum of the two weight maps at any spatial location $(x, y, z)$ strictly equals 1. The adaptively integrated feature is then computed via an element-wise weighting summation:

\begin{equation}
    \mathbf{F}_{\text{adapt}} = \mathbf{W}_{\text{local}} \odot \mathbf{F}_{\text{local}} + \mathbf{W}_{\text{context}} \odot \mathbf{F}_{\text{context}}.
\end{equation}

Finally, the ASR module employs a residual connection to ensure stable optimization. The adaptively integrated feature is passed through a convolutional layer for channel alignment before being added back to the initial input:
\begin{equation}
    \mathbf{F}_{\text{refined}} = \text{Conv}(\mathbf{F}_{\text{adapt}}) + \hat{\mathbf{F}}_{\text{fused}}.
\end{equation}

where $\text{Conv}(\cdot)$ corresponds to the standard 3D convolution for channel projection. The output $\mathbf{F}_{\text{refined}}$ is subsequently forwarded to the task head.

\section{Experiments}
\subsection{Experimental Setup}
\textbf{Datasets.} We evaluate our proposed VQSOP on the Semantic-OPV2V dataset introduced by CoHFF \cite{song2024collaborative}. This dataset is an augmented version of the large-scale collaborative perception benchmark OPV2V \cite{xu2022opv2v}, which is co-simulated by the OpenCDA \cite{xu2023opencda} and CARLA \cite{dosovitskiy2017carla} frameworks. In these simulated scenes, 2 to 7 connected vehicles exchange information, with each vehicle equipped with a 3D LiDAR sensor and four directional cameras. Because the original OPV2V dataset lacks semantic occupancy annotations, Semantic-OPV2V replays the original simulations to capture additional semantic LiDAR sweeps, thereby constructing reliable collaborative semantic occupancy supervision by aggregating the multi-agent ground truth.

\textbf{Implementation Details.} Following the experimental settings of CoHFF, the perception range is defined as $40 \times 40 \times 3.2$ m, which is discretized into a voxel grid of size $100 \times 100 \times 8$ with a voxel resolution of 0.4 m. For optimization, we employ the AdamW \cite{loshchilov2017decoupled} optimizer with a weight decay of 0.01. The learning rate undergoes a linear warmup to $2 \times 10^{-4}$ during the first 500 iterations, subsequently following a cosine annealing decay schedule. The entire framework is trained for 60 epochs with a batch size of 1 on a single NVIDIA RTX 4090 GPU. To supervise collaborative 3D occupancy prediction, the total loss consists of four terms: a voxel-wise cross-entropy loss for semantic classification, a scene-class affinity loss \cite{cao2022monoscene} to enforce structural consistency, a confidence loss for spatial mask learning, and the quantization loss.

\textbf{Evaluation Metrics.} Following previous works \cite{song2024collaborative,li2023voxformer, chen2026vision} on semantic occupancy prediction, we adopt mean Intersection-over-Union (mIoU) and class-wise Intersection-over-Union (IoU) as the primary evaluation metrics. Additionally, to assess performance in subsequent applications, we report the BEV 2D IoU for comparison with other baselines. Specifically, the predicted 3D voxel grids are projected onto the BEV plane along the height dimension to generate 2D semantic maps for evaluation.

\begin{table*}[t]
\centering
\caption{3D semantic occupancy prediction results on Semantic-OPV2V. Our proposed method achieves state-of-the-art performance under both single-agent and collaborative perception settings. The best and second-best results are highlighted in \textbf{bold} and \underline{underlined}, respectively.}
\label{tab:main_results}
\setlength{\tabcolsep}{8pt} 
\resizebox{\linewidth}{!}{
\begin{tabular}{l ccc ccc} 
\toprule
\multirow{2.5}{*}{Method} & \multicolumn{3}{c}{Single-Agent} & \multicolumn{3}{c}{Collaborative Perception} \\
\cmidrule(lr){2-4} \cmidrule(lr){5-7} 
 & CoHFF \cite{song2024collaborative} & GaussianFormer \cite{huang2024gaussianformer} & \textbf{Ours} & CoHFF \cite{song2024collaborative} & GSFusion \cite{chen2026vision} & \textbf{Ours} \\
\midrule
IoU            & 38.52 & \underline{67.76} & \textbf{70.40} & 50.46 & \underline{72.87} & \textbf{73.79} \\
mIoU           & 24.85 & \underline{29.20} & \textbf{33.61} & 34.16 & \underline{37.44} & \textbf{41.54} \\
\midrule
Building       & \textbf{21.04} & 3.84  & \underline{10.36} & \textbf{25.72} & 9.61  & \underline{11.22} \\
Fence          & \underline{20.50} & 14.10 & \textbf{30.16} & 27.83 & \underline{29.20} & \textbf{33.19} \\
Terrain        & 43.93 & \textbf{68.97} & \underline{48.26} & 48.30 & \textbf{74.51} & \underline{63.19} \\
Pole           & \textbf{31.66} & 5.94  & \underline{12.25} & \textbf{42.74} & 12.19 & \underline{20.16} \\
Road           & 55.83 & \textbf{79.37} & \underline{75.63} & 61.77 & \underline{83.05} & \textbf{87.06} \\
Side walk      & 17.31 & \textbf{70.55} & \underline{57.72} & 39.62 & \textbf{78.22} & \underline{67.40} \\
Vegetation     & \underline{14.49} & 12.54 & \textbf{23.40} & \underline{20.59} & 20.43 & \textbf{26.78} \\
Vehicles       & \textbf{58.55} & 49.25 & \underline{52.37} & \underline{63.28} & 60.49 & \textbf{63.36} \\
Wall           & \underline{33.30} & 30.79 & \textbf{35.10} & \textbf{58.27} & 36.45 & \underline{36.51} \\
Guard rail     & 1.54  & \underline{15.01} & \textbf{28.78} & 1.94  & \underline{32.50} & \textbf{54.20} \\
Traffic signs  & 0.00  & 0.00  & \textbf{13.36} & \textbf{16.33} & 8.26  & \underline{14.02} \\
Bridge         & 0.00  & 0.00  & \textbf{15.95} & 3.53  & \underline{4.35}  & \textbf{21.37} \\
\bottomrule
\end{tabular}
}
\end{table*}

\subsection{Main Results}
\textbf{3D Semantic Occupancy Prediction.} 
\Cref{tab:main_results} presents the quantitative comparison of 3D semantic occupancy prediction on the Semantic-OPV2V dataset. Compared with CoHFF \cite{song2024collaborative}, which adopts a TPV representation, 
and the Vision-Only Gaussian Splatting framework proposed by Chen et al. \cite{chen2026vision} (hereafter referred to as GSFusion), 
our VQSOP achieves notable improvements across all metrics.
Under the single-agent setting, our method reaches 70.40\% in overall IoU and 33.61\% in mIoU, significantly outperforming GaussianFormer \cite{huang2024gaussianformer} by 4.41\% in mIoU. When extended to the collaborative perception setting, our VQSOP achieves consistent IoU improvements across all semantic categories, validating the necessity and effectiveness of multi-agent information sharing. In this setting, our method achieves the highest performance with 73.79\% IoU and 41.54\% mIoU, surpassing the second-best approach (GSFusion) by 4.10\% in mIoU.
Furthermore, our method achieves the best or second-best results across all semantic categories.
Notably, class-wise analysis further reveals our method’s strength in predicting small, thin, and geometrically complex objects. 
For example, under collaborative perception, our model achieves absolute IoU improvements of 21.70\% on Guard rail and 17.02\% on Bridge compared to the second-best results. 
These substantial gains suggest that the proposed ASR module plays a critical role in modeling fine-grained structures.

\textbf{BEV Segmentation.}
\Cref{tab:bev_segmentation} presents the quantitative comparison of BEV segmentation under two collaborative settings. 
Our method consistently outperforms existing approaches in both scenarios. With 2 agents, we achieve 73.23\% IoU on Vehicle and 85.05\% on Road, surpassing the strongest baseline by 2.98\% and 2.36\%, respectively. When scaling up to 7 agents, the performance further improves to 77.48\% on Vehicle and 87.04\% on Road, maintaining clear margins over prior methods. These results indicate that the learned spatial representations remain robust and semantically consistent when applied to BEV segmentation.

\textbf{Communication Volume.} 
\Cref{tab:communication_volume} compares the communication volume (CV) and the corresponding perception performance. As observed in the baseline methods, when GSFusion reduces the number of Gaussians from 25,600 to 6,400 to lower the CV from 1.07 MB to 0.27 MB, its mIoU drops from 37.44\% to 36.02\%, indicating a strict coupling between performance and bandwidth.
In contrast, our method effectively breaks this bottleneck. It requires a CV of only 0.013 MB, which is 60$\times$ smaller than CoHFF (0.78 MB) and over 82$\times$ smaller than the high-resolution setting of GSFusion (1.07 MB). At this minimal bandwidth, our method achieves the highest performance with 73.79\% IoU and 41.54\% mIoU. This extreme communication efficiency demonstrates the effectiveness of our SAVQ module, which successfully isolates and quantizes only the essential foreground features, significantly reducing transmission costs.

\begin{table}[tb]
\centering
\caption{Quantitative results on 2D BEV semantic segmentation. We report the 2D IoU for Vehicle, Road, and Others classes under two collaborative settings: 2 agents and up to 7 agents.}
\label{tab:bev_segmentation}
\begin{tabular*}{\linewidth}{@{\extracolsep{\fill}}lcccc@{}}
\toprule
Approach & \# Agents & Vehicle & Road & Others \\
\midrule
CoBEVT \cite{xu2022cobevt}          & 2 & 46.13 & 52.41 & - \\
CoHFF \cite{song2024collaborative}           & 2 & 47.40 & 63.36 & 40.27 \\
GSFusion \cite{chen2026vision}  & 2 & 70.25 & 82.69 & 79.37 \\
Ours            & 2 & \textbf{73.23} & \textbf{85.05} & \textbf{79.42} \\
\midrule
CoBEVT \cite{xu2022cobevt}          & Up to 7 & 60.40 & 63.00 & - \\
CoHFF \cite{song2024collaborative}           & Up to 7 & 64.44 & 57.28 & 45.89 \\
GSFusion \cite{chen2026vision}  & Up to 7 & 75.30 & 84.96 & 80.19 \\
Ours            & Up to 7 & \textbf{77.48} & \textbf{87.04} & \textbf{81.10} \\
\bottomrule
\end{tabular*}
\end{table}

\begin{table}[tb]
\centering
\caption{Comparison of communication volume and performance. CV denotes the Communication Volume per agent. $\downarrow$ indicates lower is better, and $\uparrow$ indicates higher is better. Our method achieves the lowest bandwidth requirement while maintaining the highest performance.}
\label{tab:communication_volume}

\setlength{\tabcolsep}{12pt} 
\begin{tabular}{lccc}
\toprule
Method & CV (MB) $\downarrow$ & IoU $\uparrow$ & mIoU $\uparrow$ \\
\midrule
CoHFF \cite{song2024collaborative} & 0.78 & 50.46 & 34.16 \\
GSFusion  \cite{chen2026vision} (25600 Gaussians) & 1.07 & 72.87 & 37.44 \\
GSFusion  \cite{chen2026vision} (6400 Gaussians)  & 0.27 & 72.42 & 36.02 \\
Ours & \textbf{0.013} & \textbf{73.79} & \textbf{41.54} \\
\bottomrule
\end{tabular}
\end{table}

\begin{table}[tb]
\centering
\caption{Ablation study on the core components of our proposed framework. We evaluate the individual and combined effects of the SAVQ and ASR modules on communication volume and perception accuracy.}
\label{tab:ablation_components}

\setlength{\tabcolsep}{12pt} 
\begin{tabular}{cc ccc}
\toprule
SAVQ & ASR & CV (MB) $\downarrow$ & mIoU $\uparrow$ & IoU $\uparrow$ \\
\midrule
           &            & 7.32            & 40.72          & 72.48          \\
\checkmark &            & 0.013          & 40.11          & 72.06          \\
           & \checkmark & 7.32            & 41.26          & 72.88          \\
\checkmark & \checkmark & \textbf{0.013} & \textbf{41.54} & \textbf{73.79} \\
\bottomrule
\end{tabular}
\end{table}

\subsection{Ablation Study}
\textbf{Component Analysis.} 
To validate the effectiveness of our core modules, we conduct an ablation study as presented in \Cref{tab:ablation_components}. Without the SAVQ and ASR modules, the baseline achieves 72.48\% IoU and 40.72\% mIoU with a CV of 7.32 MB. When solely integrating the SAVQ module, the CV is drastically compressed from 7.32 MB to 0.013 MB. However, this extreme compression introduces a slight information loss, dropping the mIoU to 40.11\%. Conversely, applying only the ASR module improves the baseline mIoU to 41.26\% by refining the structural representations, though the CV remains at 7.32 MB. Notably, when both modules are enabled, the model achieves the best overall performance (73.79\% IoU and 41.54\% mIoU) under the minimal communication budget, demonstrating a strong synergistic effect between the two components.

\textbf{Effect of Confidence Threshold.} 
To investigate the impact of the confidence threshold $\tau$ within the Sparse-Aware selector of the SAVQ module, we conduct an ablation study as shown in \Cref{tab:threshold_ablation}. As $\tau$ increases from 0.6 to 0.8, the CV steadily decreases from 0.018 MB to 0.013 MB. Interestingly, the perception accuracy concurrently improves, peaking at 41.54\% mIoU and 73.79\% IoU. This trend suggests that a well-calibrated threshold effectively filters out task-irrelevant background noise, thereby refining the transmitted features and alleviating interference. However, when the threshold is excessively raised to 0.9, although the CV drops to its minimum of 0.012 MB, the performance suffers a noticeable decline (dropping to 41.03\% mIoU and 72.24\% IoU). This degradation indicates that an overly strict threshold inadvertently discards essential foreground structural details. Consequently, we set $\tau = 0.8$ as the default configuration to achieve the optimal balance between communication efficiency and perception accuracy.

\begin{table}[tb]
\centering
\caption{Ablation study on the confidence threshold $\tau$ in the SAVQ module. We evaluate the impact of different threshold values on both communication volume and accuracy.}
\label{tab:threshold_ablation}
\setlength{\tabcolsep}{12pt} 
\begin{tabular}{cccc}
\toprule
Threshold & CV (MB) $\downarrow$ & mIoU $\uparrow$ & IoU $\uparrow$ \\
\midrule
0.6 & 0.018 & 41.47 & 73.57 \\
0.7 & 0.015 & 41.52 & 73.66 \\
0.8 & 0.013 & \textbf{41.54} & \textbf{73.79} \\
0.9 & \textbf{0.012} & 41.03 & 72.24 \\
\bottomrule
\end{tabular}
\end{table}

\begin{table}[tb]
\centering
\caption{Ablation study on the ASR module with SAVQ enabled. We evaluate the individual contributions of the local branch, context branch, and adaptive integration mechanism.}
\label{tab:asr_components}

\setlength{\tabcolsep}{12pt}
\begin{tabular}{ccccc}
\toprule
Local & Context & Adaptive & mIoU $\uparrow$ & IoU $\uparrow$ \\
\midrule
            &            &            & 40.11 & 72.06 \\
\checkmark &            &            & 40.23 & 72.56 \\
            & \checkmark &            & 40.85 & 72.68 \\
\checkmark & \checkmark &            & 41.03 & 73.14 \\
\checkmark & \checkmark & \checkmark & \textbf{41.54} & \textbf{73.79} \\
\bottomrule
\end{tabular}
\end{table}

\textbf{Effectiveness of ASR Components.}
To validate the effectiveness of each ASR component, we conduct a fine-grained ablation study at the module level. As shown in \Cref{tab:asr_components}, the first row corresponds to the setting without ASR. When both local and context branches are enabled but adaptive integration is disabled, we use fixed equal-weight fusion. The results show that both branches improve over the setting without ASR, and the context branch achieves higher mIoU, indicating the importance of broader semantic context for feature refinement. The last two rows show the benefit of adaptive integration in better leveraging the complementarity of the two branches.

\begin{figure}[tb]
  \centering
  \includegraphics[width=\linewidth]{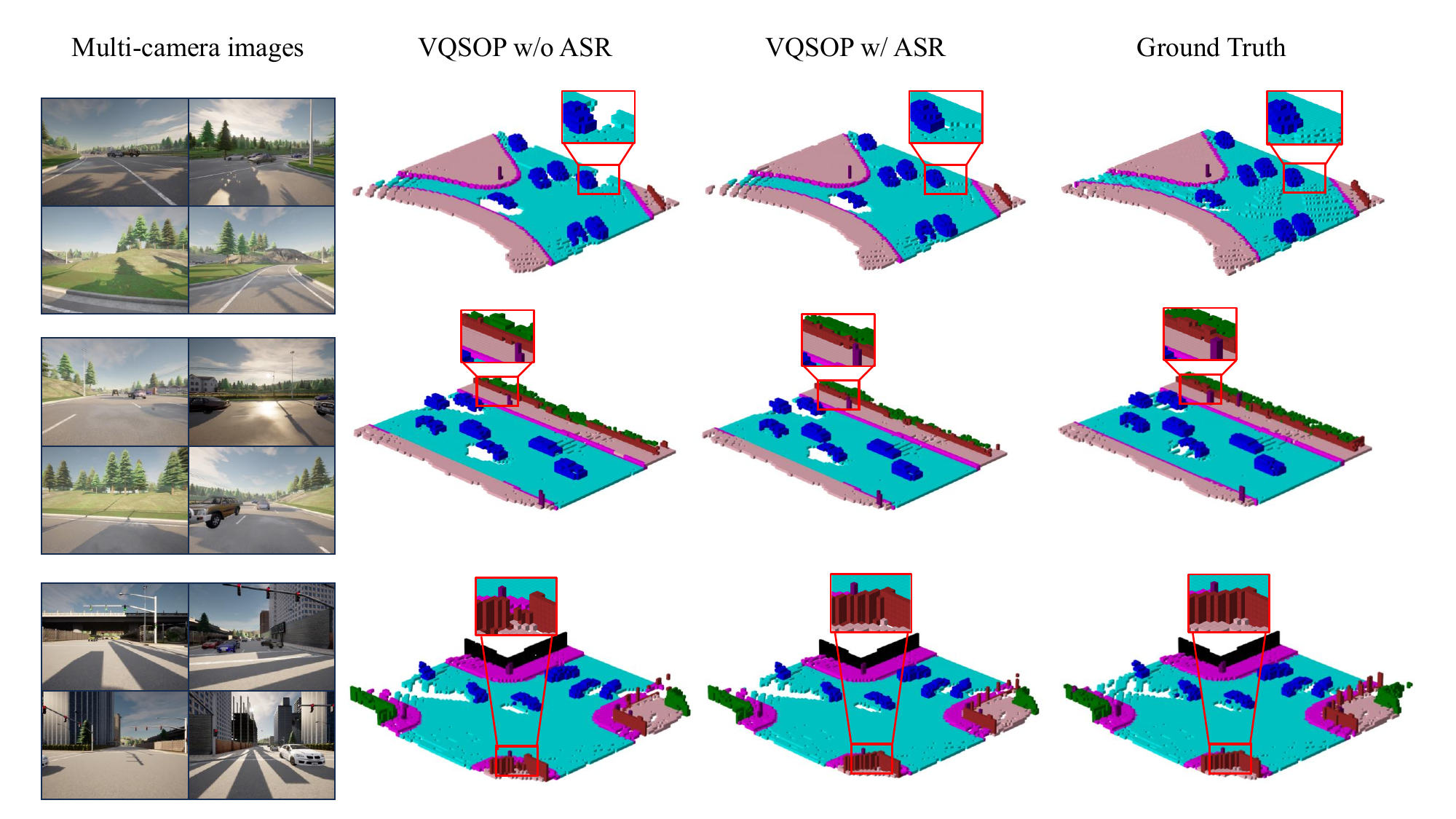}
  \caption{
    Qualitative results of 3D semantic occupancy prediction. From left to right: the input multi-camera images, predictions of VQSOP w/o ASR, our full model VQSOP w/ ASR, and the Ground Truth. The red zoomed-in regions highlight that our VQSOP equipped with ASR successfully recovers fine-grained geometric details.
  }
  \label{fig:qualitative_vis}
\end{figure}

\textbf{Visualization.} 
To provide a more intuitive understanding of the impact of the ASR module, we present qualitative comparisons in \cref{fig:qualitative_vis}. The red boxes highlight representative regions containing fine-grained structures and boundary-sensitive areas.
For VQSOP without ASR, the model produces incomplete predictions with noticeable holes in road regions and occasionally misses small objects and thin structures, such as poles and fences. These discrepancies are more evident in distant regions and geometrically complex areas.
In contrast, VQSOP with ASR aligns more closely with the ground truth. The full model better reconstructs fine-grained geometric details and preserves structural continuity, resulting in more coherent scene layouts.
These observations highlight the important role of the ASR module in refining local spatial context and improving the modeling of complex scene geometries.

\section{Conclusion}
In this paper, we proposed VQSOP, a bandwidth-efficient framework for collaborative 3D semantic occupancy prediction. The proposed approach encodes high-dimensional 3D features into compact discrete representations, significantly reducing transmission overhead while preserving essential geometric semantics. Through spatial refinement after collaborative fusion, the framework further improves structural completeness and semantic consistency. Extensive experiments on the Semantic-OPV2V dataset demonstrate that VQSOP achieves state-of-the-art performance with lower communication volume, providing an effective and practical solution for communication-efficient collaborative 3D semantic occupancy prediction.

\bibliographystyle{splncs04}
\bibliography{main}
\end{document}